\documentclass{edm_article}
\usepackage[numbers,sort&compress,square]{natbib}

\usepackage{hyperref}
\usepackage{graphicx}
\usepackage{amsmath}
\usepackage{comment}
\usepackage{float}
\usepackage[utf8]{inputenc}
\usepackage{multirow}
\usepackage{tabularx}
\usepackage{booktabs} 
\usepackage{xurl}

\begin{document}
\title{Student Answer Forecasting: Transformer-Driven Answer Choice Prediction for Language Learning}

\numberofauthors{6}
\author{
Elena Grazia Gado*\thanks{Equal contribution. Listing order is alphabetical.}, Tommaso Martorella*, Luca Zunino*, \\Paola Mejia-Domenzain, Vinitra Swamy, Jibril Frej, Tanja Käser \\
\affaddr{EPFL} \\
\email{\{elena.gado, paola.mejia, tanja.kaeser\}@epfl.ch}
}

\maketitle

\begin{abstract}
Intelligent Tutoring Systems (ITS) enhance personalized learning by predicting student answers to provide immediate and customized instruction. However, recent research has primarily focused on the correctness of the answer rather than the student's performance on specific answer choices, limiting insights into students' thought processes and potential misconceptions. To address this gap, we present \texttt{MCQStudentBert}, an answer forecasting model that leverages the capabilities of Large Language Models (LLMs) to integrate contextual understanding of students' answering history along with the text of the questions and answers. By predicting the specific answer choices students are likely to make, practitioners can easily extend the model to new answer choices or remove answer choices for the same multiple-choice question (MCQ) without retraining the model. In particular, we compare MLP, LSTM, BERT, and Mistral 7B architectures to generate embeddings from students' past interactions, which are then incorporated into a finetuned BERT's answer-forecasting mechanism. We apply our pipeline to a dataset of language learning MCQ, gathered from an ITS with over 10,000 students to explore the predictive accuracy of \texttt{MCQStudentBert}, which incorporates student interaction patterns, in comparison to correct answer prediction and traditional mastery-learning feature-based approaches. This work opens the door to more personalized content, modularization, and granular support.
\end{abstract}


\section{Introduction}
\label{intro}
\begin{figure*}[ht!]
  \centering
  \includegraphics[width=0.8\textwidth]{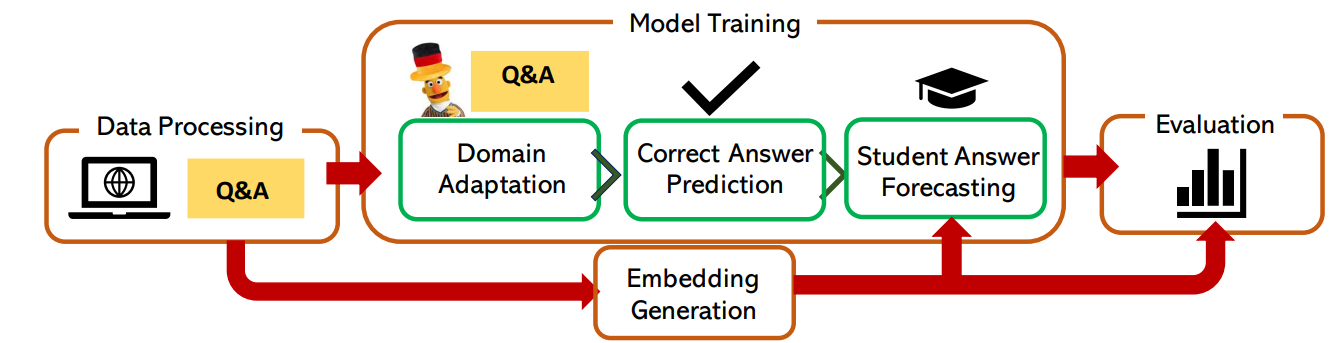}
  \caption{We present a four-stage pipeline for answer forecasting integrating student history: 1) we preprocess ITS data from \textit{Lernnavi}, 2) we train student embeddings from several models (MLP, LSTM, Mistral 7B, BERT), 3) we use the ITS data and student embeddings to train several models (\texttt{LernnaviBERT} for Domain Adaptation, \texttt{MCQBert} for Correct Answer Prediction, and \texttt{MCQStudentBert} for Student Answer Forecasting) and 4) we evaluate the models using qualitative and quantitative analyses (accuracy, F1 Score, and MCC).}
   \Description{The image is a flowchart depicting the answer forecasting pipeline which includes several stages. The first stage is "Data Processing," represented by a laptop icon, which prepares Q\&A data for the model. Next is the "Model Training," which involves three steps: "Domain Adaptation", "Correct Answer Prediction", and "Student Answer Forecasting". The last stage is the "Evaluation" stage, depicted by a bar chart, where the model's performance is assessed. The flowchart includes arrows indicating the flow of data between these stages.}
  \label{fig:pipeline}
\end{figure*}

Intelligent tutoring systems (ITS) are powerful educational tools that personalize the student's learning experience through adaptive content \cite{heffernan2014assistments,abdelrahman2023knowledge,kaser2013design}. Within these systems, the ability to predict student answers plays an important role in tailoring the educational content to the student's level of understanding, knowledge gaps, and learning pace \cite{anderson1995cognitive}. 

There is a large body of research modeling students' learning \cite{heffernan2014assistments,abdelrahman2023knowledge, kaser2013design,anderson1995cognitive,shah2020explainable,corbett1994knowledge,piech2015deep,nakagawa2019graph,Sarsa_Leinonen_Hellas_2022}. This effort encompasses the development of probabilistic frameworks such as Bayesian Knowledge Tracing \cite{corbett1994knowledge} and Dynamic Bayesian Networks \cite{kaser2013design}, as well as deep learning approaches like Deep Knowledge Tracing \cite{piech2015deep} and Graph-based Knowledge Tracing \cite{nakagawa2019graph}. Other educational data mining (EDM) approaches have developed statistical models such as Learning Factor Analysis \cite{cen2006learning} and Performance Factor Analysis \cite{pavlik2009performance} to predict the probability of correct student responses. Additionally, the EDM community has studied the implementation of Machine Learning (ML) classifiers to predict learning outcomes such as quiz answers \cite{lincke2021performance}.

Despite these advancements, the focus has predominantly been on predicting whether a student's answer will be correct or incorrect \cite{heffernan2014assistments,abdelrahman2023knowledge,kaser2013design,anderson1995cognitive,corbett1994knowledge,piech2015deep,lincke2021performance}, rather than forecasting the specific answer the student would provide. This could enrich the understanding of the student's acquired knowledge. Thus, enabling the development of more personalized content and hints \cite{wang2020diagnostic}.

Several works have tackled the challenge of analyzing Multiple Choice Questions (MCQs) \cite{which_choose, what_does_BERT,results_insights}. For example, \cite{lincke2021performance} and \cite{shinahara2020quality} incorporated temporal features, user history features, and subject features to train ML classifiers, such as XGBoost, to predict question quality. Additionally, \cite{zhang2020} utilized a transformer model to fuse metadata and performance features for a multiclass classification task. Another approach by \cite{ghoshoption} extended Binary Knowledge Tracing using a BiLSTM with DAS3H features \cite{choffin2019das3h} and attention mechanisms. Similarly, \cite{which_choose} proposed the Order-aware Cognitive Diagnosis (OCD) model to predict students' answers by considering question order effects, without focusing on the question or answer text. However, a common limitation in these studies is the lack of attention to the contextual richness in the text of questions and answers, which could significantly influence human cognition and decision-making processes.

In this regard, Large Language Models (LLMs) could be leveraged to incorporate textual context into predictive models \cite{hayat_gaied_23,TAL_ML,nguyen-23-synthesizing}. For example, \cite{hayat_gaied_23} used LLMs fine-tuned with personalization and contextualization to enhance early forecasting of student performance in courses. Moreover, \cite{TAL_ML} proposed a transformer-based knowledge tracing model using BERT to capture the sequential knowledge states by randomly masking labels from the students' answer sequence.  

While LLMs offer a promising solution to account for the content and context of questions and answers, their application on student answer forecasting remains underexplored. To forecast student answers, the inputs of the question context, granular answer choices, and individual learning history becomes even more relevant to the model than for general question answering tasks \cite{shishir2021,robinson2022leveraging}. 

To address this gap, we introduce a novel student answer forecasting pipeline that leverages LLMs to understand the content and context of the question and answer and the students' history. We first compare four architectures (MLP, LSTM, BERT, Mistral 7B) to compute student embeddings using a student's previous answering history. Then, we incorporate the student embedding into the question-answering prediction, using a finetuned BERT architecture. We focus on language learning MCQs from a real-world ITS used by 10499 students consisting of 237 unique questions to answer the following research questions: \textbf{(RQ1)} How can we design a performant embedding for student interactions in German? \textbf{(RQ2)} How can we integrate these student interaction embeddings to improve the performance of an answer forecasting model?

This work contributes a modeling pipeline for question-answer forecasting that 1) integrates student history into a transformer model and 2) focuses on answer choice forecasting instead of correct answer forecasting. Unlike other answer forecasting models, \textit{answer choice forecasting} allows for independent modularization of answer choices, enabling an educator to simply add a fifth answer choice for an original four-answer MCQ question without retraining the model. Importantly, we contribute to the literature in German EDM, presenting a case study from over ten thousand students from a real-world ITS in a language that is not often researched and therefore accompanied by several biases due to data and model underrepresentation \cite{wambsganss2022bias}. We only use open-source models (including the recent Mistral 7B) and not API-based services, enabling learning platforms to host their pipeline and data entirely on their servers. Our code and models are provided open source at \url{https://github.com/epfl-ml4ed/answer-forecasting} and \url{https://go.epfl.ch/hf-answer-forecasting}.
\section{Methodology}
\label{sec:methods}
The student answer forecasting pipeline depicted in Figure \ref{fig:pipeline} is based on students interactions with MCQs in an ITS named \emph{Lernnavi}. The pipeline predicts the likelihood of a student selecting a particular MCQ answer, based on the question and answer text and a student embeddings generated from their historical interaction data. In this section, we describe each step of the pipeline.

\subsection{Data Processing}
\vspace{1mm} \noindent \textbf{Learning Context}. We focus our analyses on data collected from \textit{Lernnavi}, an ITS for high-school students. \textit{Lernnavi} offers adaptive learning and testing sessions in mathematics and language learning.

\noindent \textbf{Dataset}. The dataset is characterized by the following three data representations: 1) user-generated interactions, also referred to as ``transactions`` ($I_u^s = {i_1, \ldots, i_K}$ for each user $u$), 2) \textit{Lernnavi} questions also referred to as ``documents'', representing the associated questions ($\mathbb{Q}^s$), answer choices ($\mathbb{C}^s$), and textual pages provided to students, and 3) the taxonomy of topics ($\mathbb{T}$) shown in the German and Math dashboards. We only consider ``documents'' regarding German MCQs with at least one transaction from a user. After filtering, the dataset is composed of 237 unique questions and 138,149 transactions.
Moreover, the dataset consists of 10,499 users with at least one transaction for German MCQs. The median number of MCQ answers from learners is 7 with some learners that answered up to 311 questions (including multiple trials for the same question).

\subsection{Problem Formulation}
We analyze \emph{users} $U^s \subset \mathbb{U}$ engaging in \emph{learning sessions} $s$ within \textit{Lernnavi}'s $\mathbb{S}$ offerings, focusing on sessions $S = {s_1, \ldots, s_{M^S}}$, each a unique iteration within a broader topic $t \subset \mathbb{T}$. These sessions are characterized by their interactive quizzes, sourced from a question bank $\mathbb{Q}^s$ and designed to assess user knowledge through multiple-choice formats. Interactions in these sessions are represented as $I_u^s = {i_1, \ldots, i_K}$ for each user $u$, involve selections ($c$) from the provided answer options for each question $q$. 

These interactions are timestamped and detailed to capture the essence of user engagement and learning behavior. To evaluate user trajectories, we introduce \emph{binary metrics} for answer choices, $\mathbb{C}^s = {c_{q_1}, \ldots, c_{q_{|Q^s|}}}$, allowing for an in-depth analysis of user response selection. This design is to enable the multi-response setting for question $q$, which can either have one correct answer choice $c_{q_i}$ or multiple correct answer choices $\mathbb{C}_{q}$, of which user $u$ chooses one answer $c_{q_u}$ or multiple $\mathbb{C}_{q_u}$.

The answer-forecasting prediction task is to predict for a given user $u$ with past interaction history $I_u^s$, which answer choices $\mathbb{C}_{q}$ are most likely to be chosen by the student.

\subsection{Embedding Generation}
\label{StudentEmbedding}

To create student embeddings for a prediction model, we explored four strategies to make a total of 22 different embeddings: one using the MLP model, 16 using the LSTM models, one using \texttt{LernnaviBERT}, and four using \texttt{Mistral 7B Instruct} models: \vspace{1mm}\\
\noindent {\textbf{{MLP Autoencoder Embedding:}}} We utilized a Multilayer Perceptron (MLP) autoencoder with specific architecture and feature engineering to encode students' previous performance, resulting in a size of 11907. \vspace{1mm} \\
\noindent {\textbf{{LSTM Autoencoder Embedding:}}} We considered four stacked LSTM configurations with varying sequence lengths and number of layers to balance computational complexity and richness of student interaction history. \vspace{1mm} \\
\noindent {\textbf{{LernnaviBERT Embedding:}}} We created embeddings using a finetuned German BERT base model\footnote{\small{https://huggingface.co/google-bert/bert-base-german-cased}} to MCQ-specific language, \texttt{LernnaviBERT}, with a sequence length of 10 and mean pooling strategy. \vspace{1mm} \\ 
\noindent {\textbf{{Mistral 7B Instruct Embedding:}}} We used \texttt{Mistral 7B Instruct} to generate embeddings with sequence lengths of 10, 20, 30, and 40 with mean pooling at the penultimate layer.

\begin{figure}[h]
  \centering
  \includegraphics[width=0.98\columnwidth]{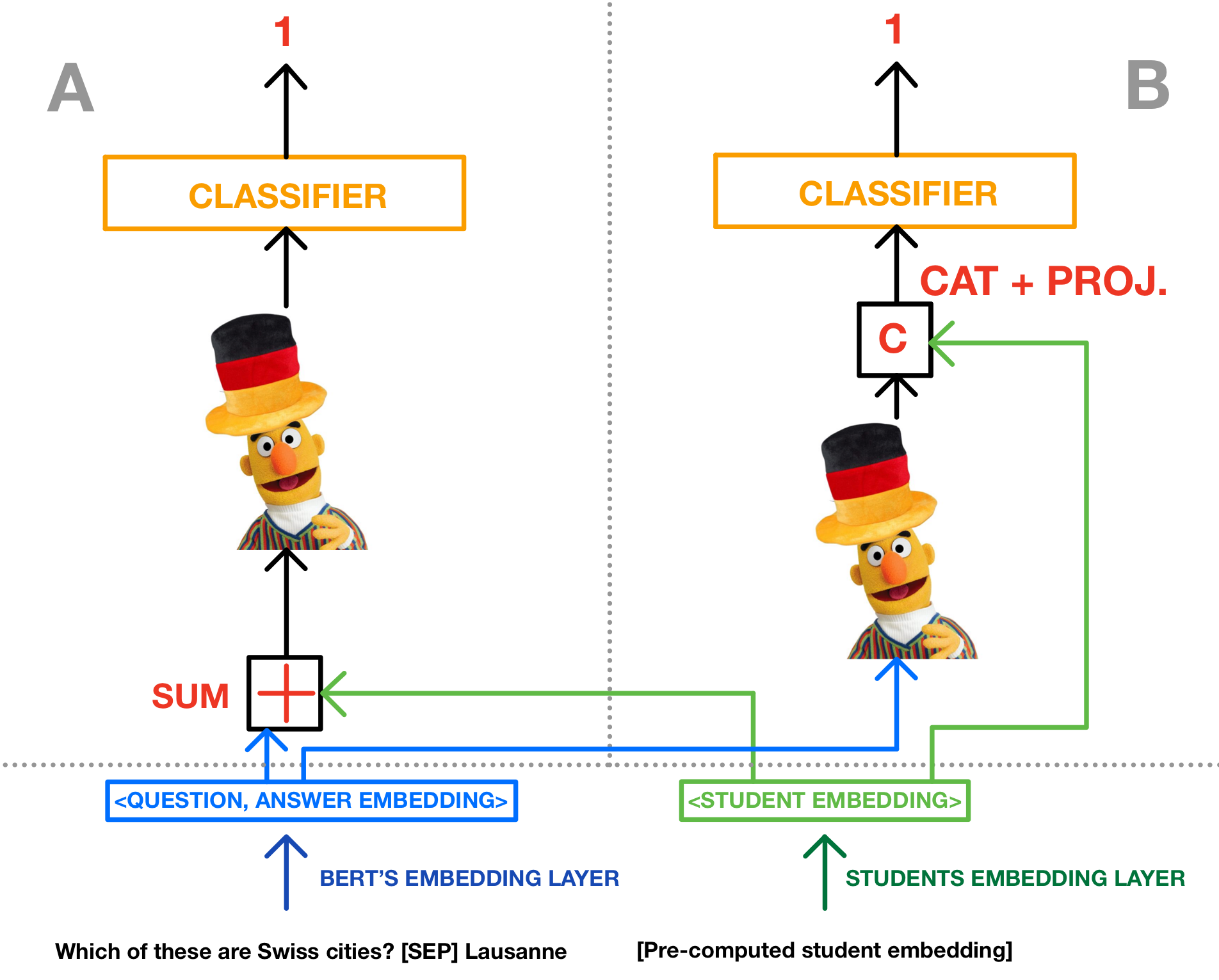}
  \caption{\textbf{\texttt{MCQStudentBertSum} (A) and \texttt{MCQStudentBertCat} (B) architectures.} In \texttt{MCQStudentBertSum}, the student embeddings are summed with \texttt{LernnaviBERT} question embeddings at the input, before being passed to the \texttt{MCQBert} model and classification layer. In \texttt{MCQStudentBertCat}, MCQ embeddings are generated with \texttt{LernnaviBERT}, then passed to the \texttt{MCQBERT} model and concatenated with the student embeddings just before the classification layer. German BERT image taken from \url{https://www.deepset.ai/german-bert}}
  \Description{The image shows two architectures for multiple-choice question (MCQ) models: MCQStudentBertSum (A) and MCQStudentBertCat (B). Both architectures include a diagram of a puppet figure, with arrows indicating the flow of embeddings. }
  \label{fig:mcq-cat-sum}
\end{figure}
\vspace{-1mm}
\subsection{MCQStudentBert: Answer Forecasting} 
\label{sec:mcqstudent}

We initially train an \texttt{MCQBert} for the classification of correct/incorrect MCQ answers. Learning the correct answers across all MCQs is necessary to ensure that any failure in predicting a student's response did not arise from a lack of knowledge of the correct answer. Appendix \ref{sec:mcqbert} details the training and evaluation strategy.

We then extend \texttt{MCQBert} to include the students' history and predict students' responses to MCQs. Inputs for this task include the text of the MCQs and supplementary student-specific data encapsulated within embeddings. The objective has changed; rather than pinpointing the correct answers from available options, the emphasis is on predicting the actual responses provided by students. This task continues to be treated as a binary classification problem. Details are included in Appendix \ref{AppendixStudent}.

We explore the two models in Figure \ref{fig:mcq-cat-sum}, differing in their handling of student embeddings for integrating student information into the prediction process: \texttt{MCQStudentBertCat} where the inputs are concatenated before the classification layer and \texttt{MCQStudentBertSum}\footnote{\small \texttt{MCQStudentBertCat} and \texttt{MCQStudentBertSum} are available at \url{https://go.epfl.ch/hf-answer-forecasting}} where the embeddings are summed at the input. Both models are based on \texttt{MCQBert}, finetuned in the previous phase of the pipeline. These variants are augmented with a classifier head, comprising two linear layers with a ReLU activation function, to predict the likelihood of each potential answer being chosen by a student. 
Inset B of Figure \ref{fig:mcq-cat-sum} illustrates \texttt{MCQStudentBertCat} strategy. The concatenation strategy draws inspiration from context-aware embeddings \cite{li2020enhancing,vaswani2023attention}, where additional features (like user or product embeddings) are appended before the final classification layer to provide context. For this purpose, in our context, the student embeddings are first transformed using a linear layer to match the \texttt{MCQBert}'s hidden size. These transformed embeddings are then concatenated with the output of the \texttt{MCQBert} model which is the representation of the first token [CLS] token. This approach leaves the \texttt{MCQBert} processing unchanged and appends additional information right before the final decision-making process (e.g., classification). It allows the classification model to consider both the processed input representation and the student-specific information distinctly.

In contrast, \texttt{MCQStudentBertSum}, depicted in Figure \ref{fig:mcq-cat-sum} (inset A), integrates the student embeddings directly into the input embeddings of the \texttt{MCQBert} model. This approach is similar to multimodal learning for LLMs to create combined embeddings that represent both modalities at the input level (e.g. to create visual-semantic embeddings) \cite{chen2021learning}. Specifically, the student embeddings are first transformed to match the dimensionality of the \texttt{MCQBert} input embeddings using a linear layer. These transformed student embeddings are then summed with the original input embeddings. This approach alters the initial representation that the \texttt{MCQBert} model processes. The student embeddings can be seen as providing an initial bias or modification to the input embeddings, potentially allowing the model to adapt more specifically to characteristics represented by the student embeddings.
\section{Experimental Evaluation}
\label{results}
We finetune \texttt{LernnaviBert} to predict the correct MCQ answer, resulting in \texttt{MCQBert}. Next, we integrate the student embeddings (\textbf{RQ1}) to forecast student answers to produce variations of \texttt{MCQStudentBert} (\textbf{RQ2}). The experimental evaluation of \texttt{MCQBert} for correct answer prediction can be found in Section \ref{sec:correct-answer}

\begin{table*}[t]
\centering
\caption{\texttt{MCQStudentBert} Embedding Comparisons and Performance. For two models \texttt{MCQStudentBertCat} and \texttt{MCQStudentBertSum}, we evaluate the performance of different student embeddings (MLP, LSTM, \texttt{LernnaviBERT}, and Mistral 7B) against a baseline dummy classifier and \texttt{MCQBert}. We report MCC, F1 Score and Accuracy, with the highest values indicating the most performant embeddings (Appendix \ref{AppendixStudent}). * represents the best model determined by a hyperparameter search.}
\begin{tabularx}{\textwidth}{@{}Xlllllll@{}}
\multicolumn{2}{c}{} & \multicolumn{6}{c}{\textbf{Embedding}} \\  \addlinespace[0.2em] \cmidrule{3-8}  \addlinespace[0.4em]
\multicolumn{2}{c}{} & Dummy & \texttt{MCQBert} & MLP & LSTM$^*$ & \texttt{LernnaviBERT} & Mistral 7B$^*$ \\ \midrule \addlinespace[0.2em]
\multirow{3}{*}{\texttt{MCQStudentBertCat}} & MCC       & 0 & 0.518 & 0.557 & 0.567 & 0.575 & \textbf{0.579} \\
                                   & F1 Score  & 0.305 & 0.740 & 0.772 & 0.777 & 0.780 & \textbf{0.782} \\ 
                                    & Accuracy & 0.590 & 0.771 & 0.785 & 0.790 & 0.795 & \textbf{0.797} \\ \midrule
                                 \addlinespace[0.2em]
\multirow{3}{*}{\texttt{MCQStudentBertSum}}  & MCC       & 0 & 0.518 & 0.552 & 0.564 & 0.568 & \textbf{0.569} \\  
                                   & F1 Score  & 0.305 & 0.740 & 0.767 & 0.774 & 0.777 & \textbf{0.778} \\ 
                                   & Accuracy & 0.590 & 0.771 & 0.785 & \textbf{0.790} & 0.789 & 0.789 \\
                                   \bottomrule
\end{tabularx}

\label{tab:results}
\end{table*}

\label{bert_results}

We evaluate the different embedding strategies and ways of integration to predict student responses to MCQ. A key difference from \texttt{MCQBert} is the incorporation of student embeddings, enabling models to use contextual information.

\textbf{Embedding Performance. }
The training of the MLP autoencoder yielded a mean validation loss of $1.3e^{-7}$. Further analysis highlighted a discrepancy in the norm-2 distance between the input and output vectors across training, validation, and test datasets, with an average input norm-2 ($\lVert input \rVert_2$:) of 1.31 and an average discrepancy norm-2 ($\lVert input - output \rVert_2$) of 0.03. The LSTM models yielded a mean validation loss of $1.3e^{-2}$ with a mean input norm of 13.59 and an average reconstruction norm of 10.94. 

We did not find a trend in the number of hidden layers for the LSTM. For subsequent analyses, the single-layer LSTMs are used. After tuning, for both  \texttt{Mistral 7B} and LSTMs, the best sequence length was 20.

\begin{figure}[t]
    \centering
    \includegraphics[width=0.8\columnwidth]{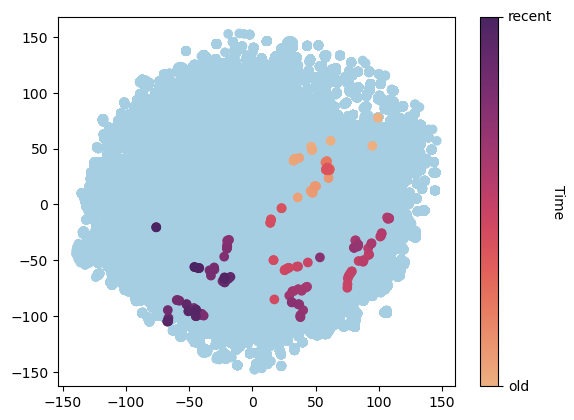}
    \caption{t-SNE visualization of the \texttt{LernnaviBERT} embedding space. Highlighted is the evolution of a single student's embedding through time.}
    \Description{The image shows a light blue circle made out of smaller blue points. The t-SNE visualization embedding of a single student are shown as colored points that stand out from the small blue circles. }
    \label{fig:tsne}
\end{figure}

To examine the \texttt{LernnaviBERT} embeddings in detail, Figure \ref{fig:tsne} visualizes the embedding space using t-SNE. Each point represents a student-embedding at a given time. Students are represented by multiple points, reflecting the evolution of their embeddings as they respond to successive MCQs. In Figure \ref{fig:tsne}, the trajectory of embeddings from an individual student is accentuated. The temporal aspect of these embeddings is depicted through a color gradient, transitioning from lighter shades for initial interactions to darker shades for more recent activity. The highlighted trajectory suggests a discernible shift in the student's embedding space over time from the upper right to the lower left corner.

\textbf{Predictive Model Performance. }
A total of 20 models were trained, incorporating 10 distinct embeddings and 2 integration strategies, over three epochs\footnote{\small \label{fn:full-results}Full results are available at \url{https://go.epfl.ch/mcq-results}}. The models were evaluated on a hold-out test set consisting of MCQs previously encountered by the model, but not in the context of the specific student being assessed. 

The optimal performance for all models was recorded in either the second (followed by a marginal decline in the third epoch) or third epoch. Table \ref{tab:results} presents the results from the best epoch, showcasing both integration strategies (concatenation and addition) across four embedding types (MLP, LSTM, \texttt{LernnaviBERT}, \texttt{Mistral 7B}), compared to a baseline Dummy Classifier and \texttt{MCQBert} (no embedding). For brevity, only the results from the best-performing LSTM, with 1 hidden layer and a sequence length of 20, and the highest-achieving \texttt{Mistral 7B} model, with a sequence length of 20, are displayed\footref{fn:full-results}.

\noindent {\textit{Integration Strategies}}.
As seen in Table \ref{tab:results}, the concatenation strategy (\texttt{MCQStudentBertCat}) generally yields slightly better results compared to the summation strategy (\texttt{MCQStudent BertSum}), particularly noticeable with \texttt{LernnaviBERT}. 

\noindent {\textit{Embeddings}}. All embedding strategies show substantial improvements over the Dummy Classifier across all metrics. \texttt{Mistral 7B} is the best-performing embedding for both integration strategies (\texttt{MCQStudentBertCat}, \texttt{MCQStudentBertSum}). When applied to \texttt{MCQStudentBertCat}, the \texttt{Mistral 7B} embedding shows a notable increase in performance metrics: an improvement of 0.579 in MCC, 0.477 in F1 score, and 0.207 in accuracy compared to the Dummy Classifier. For the second baseline (\texttt{MCQBert}), the \texttt{Mistral 7B} embedding showed a 12\% improvement.

Consistency is observed in the performance ranking of embeddings between the two integration strategies. In the \texttt{MCQStudentBertCat} configuration, \texttt{LernnaviBERT} ranks second with an $MCC=0.575$, followed by the LSTM autoencoder with $MCC=0.567$, and the MLP autoencoder trailing with $MCC=0.557$. Similarly, for the \texttt{MCQStudentBertSum} strategy, the \texttt{LernnaviBERT} embedding is in the second position $MCC=0.575$ followed by the LSTM autoencoder with an $MCC=0.564$, while the MLP autoencoder remains the least effective, with an $MCC=0.552$.

The performance differentials between embeddings are marginal. For instance, within the \texttt{MCQStudentBertCat} framework \texttt{Mistral 7B}, exhibits a modest 4\% improvement in MCC over the MLP autoencoder. Similarly, in the \texttt{MCQStudentBertSum} framework, the margin is 3\%.
\section{Discussion and Conclusion}
Our goal is to enhance the predictive capabilities of ITS by developing embeddings that capture student interactions (\textbf{RQ1}) and integrating them into an answer forecasting model (\textbf{RQ2}) to improve performance and personalization.

We explored various methods of encoding students' interactions with the ITS including using autoencoders with MLP and LSTM architectures, and LLMs including \texttt{LernnaviBERT} model and \texttt{Mistral 7B} (\textbf{RQ1}). Our findings revealed that the \texttt{Mistral 7B} embedding emerged as the best-performing method for our use-case, demonstrating a 12\% performance enhancement relative to \texttt{MQCBert} (baseline with no embedding) and a 4\% improvement over the least effective embedding, the MLP autoencoder. The performance of \texttt{Mistral 7B}, closely followed by \texttt{LernnaviBERT}, LSTM, and MLP autoencoder can be likely attributed to the inherent capabilities of each embedding approach in capturing and representing student interactions. For example, \texttt{Mistral 7B} has a sliding window attention mechanism, facilitating a deeper understanding of contextual relationships within student data \cite{jiang2023mistral}. This model's efficacy is further augmented by its fine-tuning on instructional datasets, potentially enhancing its proficiency in interpreting question-answer pairs. Moreover, \texttt{LernnaviBERT} also seemed to capture the contextual information of educational interactions effectively, ranking it closely behind \texttt{Mistral 7B}. The slight difference in performance between these two models may be due to \texttt{Mistral 7B}'s more advanced mechanisms for handling long sequences and its ability to incorporate broader contextual information. Notably, the optimal sequence length for \texttt{Mistral 7B} was identified as 20, whereas the \texttt{LernnaviBERT} model was constrained to sequence lengths of 10 due to its context limitations. This is further supported by the \texttt{Mistral 7B} embedding visualizations that show a more discernible trend with sequence lengths greater than 10. The LSTM autoencoder underperforms compared to transformer-based models because it prioritizes temporal dynamics over deep contextual understanding. While its sequential processing is good for capturing learning progression, it may not handle complex language structures as well as transformer models. The low performance scores of the MLP autoencoder can be attributed to its simpler architecture, which may not capture complex language-based information and temporal dynamics effectively.

To study the integration of student interaction embeddings into an answer forecasting model (\textbf{RQ2}), we used two approaches: \texttt{MCQStudentBertCat}, which concatenates student embeddings with model outputs before classification, and \texttt{MCQStudentBertSum}, which sums the embeddings at the input stage. The superior performance of the \texttt{MCQStudentBertCat} model compared to the \texttt{MCQStudentBertSum} model could be attributed to its ability to maintain a clear separation between the question-answer information and the student-specific embeddings, promoting distinct utilization of both sources of information in the prediction process. Future research could further explore the optimization of such integration techniques, potentially investigating the impact of varying the point of concatenation.

\vspace{-1mm}
One limitation of our study is the interpretability of the embeddings generated by the models we explored, such as \texttt{LernnaviBERT} and \texttt{Mistral 7B}. Despite their effectiveness, there is a significant gap in our understanding of the underlying features and feature patterns encapsulated by these embeddings, hindering our ability to comprehend their effectiveness.
The generalizability of our findings is limited by the study's execution within a single context and the lack of publicly available datasets comparable in richness to \textit{Lernnavi}. However, our study aligns with and contributes to the growing body of research in answer forecasting by incorporating student history into predictive models. We aim to introduce a novel approach that can be replicated by the EDM community in different ITS and contexts, enabling a better understanding of the generalizability of our findings and fostering advancements in personalized instruction.

\vspace{-1mm}
In conclusion, we introduce \texttt{MCQStudentBert}, a model for student answer forecasting that leverages LLMs to integrate the contextual understanding of question and answer texts with students' historical interactions. Our work contributes to the field of EDM in the German language context, where such studies are scarce, and promotes the use of open-source models, facilitating wider adoption and adaptation within the EDM community. Furthermore, our model's utility extends to ITS, where it can be employed to tailor potential answers for individual learners and give hints dynamically. From the educator's and developers' perspective, it is possible to modify or augment the answer choices without necessitating a complete retraining of the model. This feature could allow seamless updates and expansions to the answer sets in response to evolving pedagogical requirements or teacher/student feedback.

\noindent \textbf{Acknowledgements}. This project was substantially funded by the Swiss State Secretariat for Education, Research and Innovation (SERI) and the Swiss Canton of St. Gallen.

\bibliographystyle{unsrt}
\bibliography{literature}

\clearpage
\pagenumbering{Roman}
\section{Appendix}
\label{appendixA}
This appendix includes additional figures and material to complement the main section of the paper.

\begin{figure}[!hbpt]
  \centering
  \includegraphics[width=1\columnwidth]{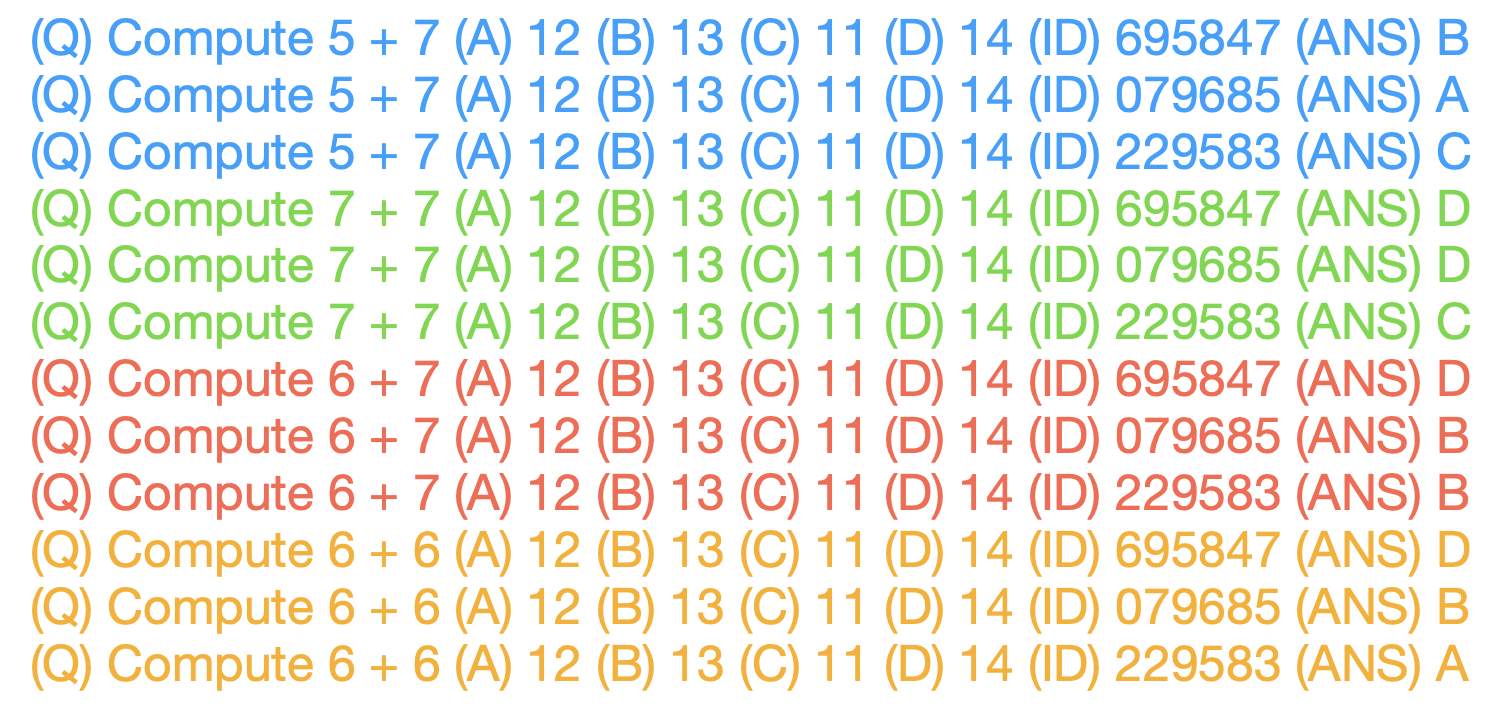}
  \caption{Example of dataset to illustrate the dataset split into a training and a validation/test dataset. For clarity, each individual MCQ is characterized by the same colour.}
  \Description{A dataset example showing multiple-choice questions in four colors: blue, green, red, and orange. Each question includes a computation problem (e.g., Compute 5 + 7) with four answer choices labeled A, B, C, and D, an ID number, and the selected answer.}
  \label{fig:1}
\end{figure}

\begin{figure}[!hbpt]
  \centering
  \includegraphics[width=1\columnwidth]{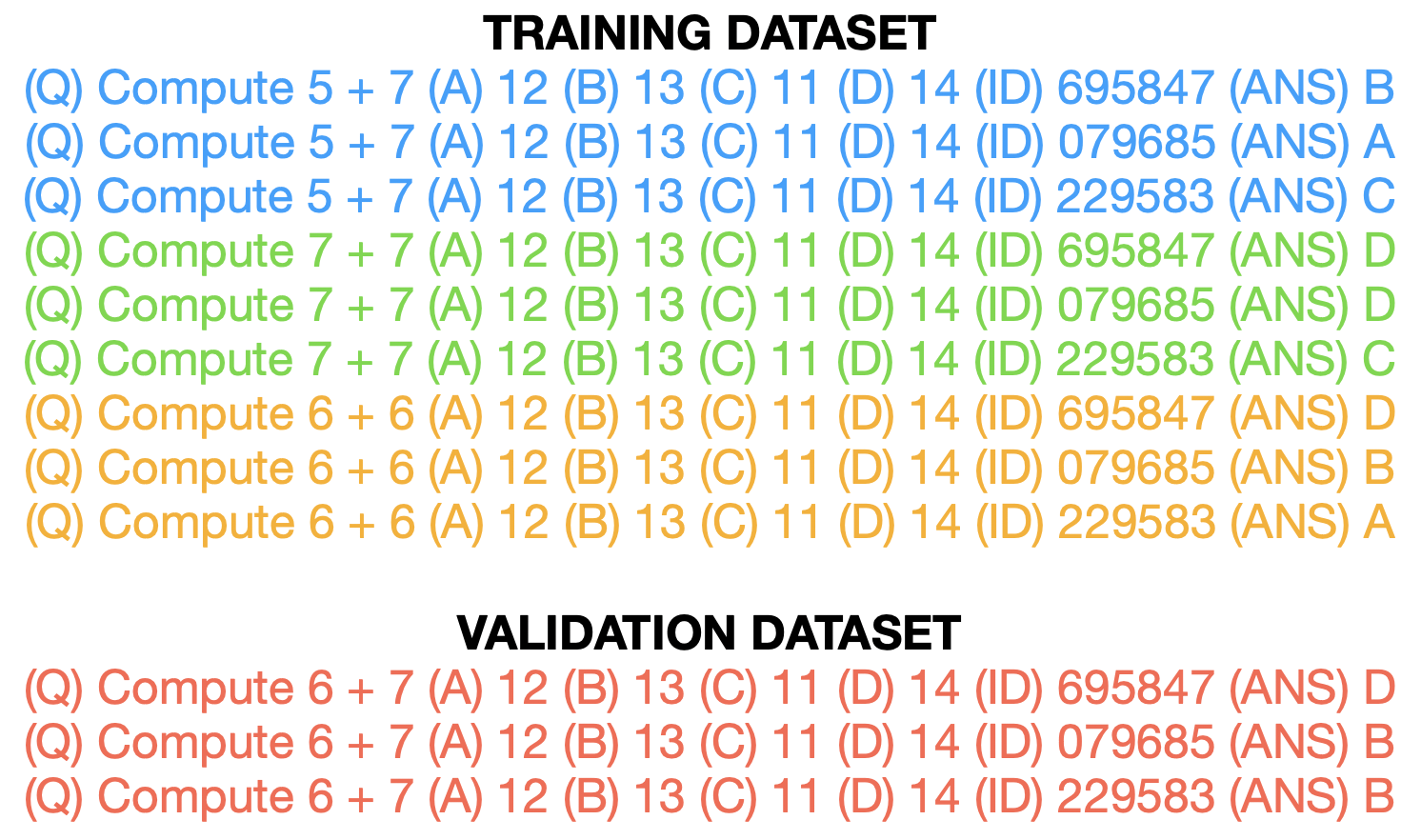}
  \caption{Training and validation/test datasets obtained by splitting the dataset using the second method presented in the report. Different instances of the same MCQ can be in only one of the two datasets.}
  \Description{Dataset example split into training and validation datasets. The training dataset includes questions in blue, green, and orange, asking to compute sums (e.g., Compute 5 + 7, Compute 7 + 7, Compute 6 + 6), each with four answer choices and an ID number. The validation dataset, in red, contains different instances of a question asking to compute 6 + 7, also with four answer choices and an ID number.}
  \label{fig:3}
\end{figure}

\begin{table}[!hbpt]
    \centering
    \caption{\texttt{MCQBert} performance in comparison with a baseline model (Dummy Classifier) that predicts the majority class (0) for each answer choice.}
    \label{tab:results-exp1}
    \vspace{10pt}
    \begin{tabular}{cccc}
    \toprule
         &  \textbf{MCC}  & \textbf{F1 Score} & \textbf{Accuracy}  \\ \addlinespace[0.2em]
         \hline
         Dummy Classifier & 0 & 0.292 & 0.605 \\ \addlinespace[0.3em]
         \texttt{MCQBert} & 0.472 & 0.702 &  0.740 \\
         \bottomrule
    \end{tabular}
\end{table}

\begin{figure}[hbpt]
  \centering
  \includegraphics[width=1\columnwidth]{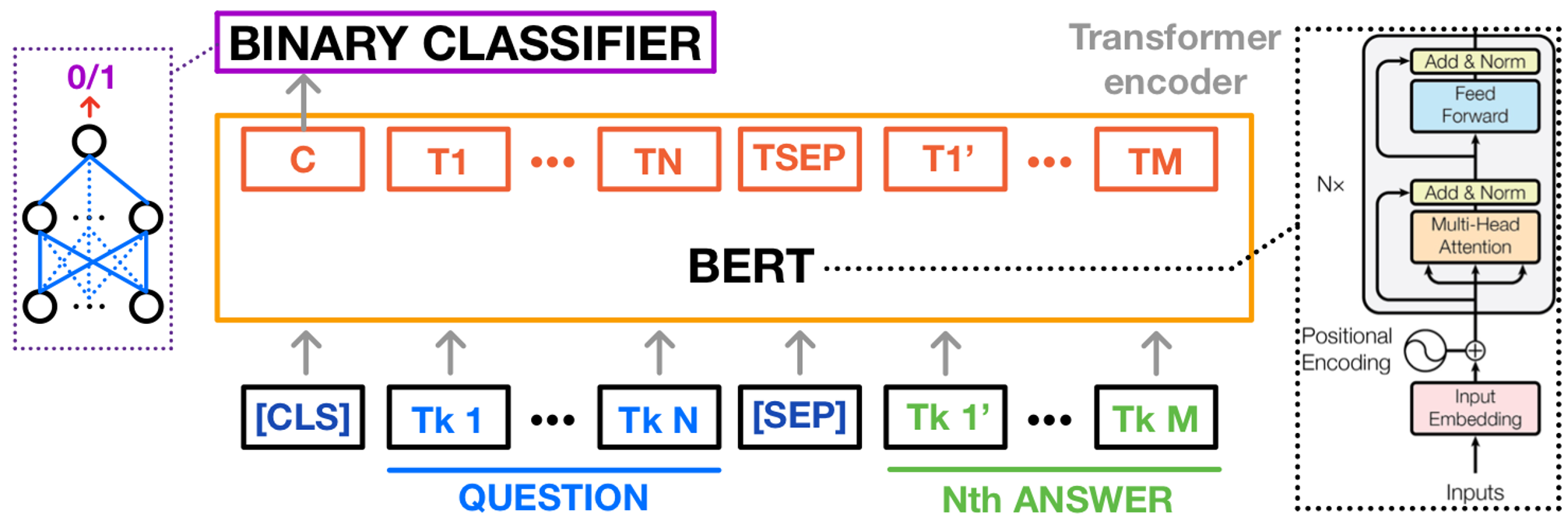}
  \caption{\textbf{\texttt{MCQBERT} Architecture}. The \texttt{MCQBERT} model for correct answer prediction involves a binary classifier and a finetuned BERT architecture predicting answers in sequence. The inset showing the Transformer encoder is taken from \cite{vaswani2023attention}.}
  \Description{At the top, a binary classifier receives input from the [CLS] token of a BERT model. The BERT model processes input sequences including a classification token [CLS], question tokens, a separator token [SEP], and answer tokens. The BERT model consists of a transformer encoder, depicted as layers of multi-head attention and feed-forward neural networks with positional encoding. An inset on the right illustrates the internal structure of the transformer encoder.}
  \label{fig:arch-bert}
\end{figure}

\begin{figure*}[htbp]
    \centering
    \includegraphics[clip, trim=5.7cm 18.9cm 0.4cm 0.2cm, width=0.98\textwidth]{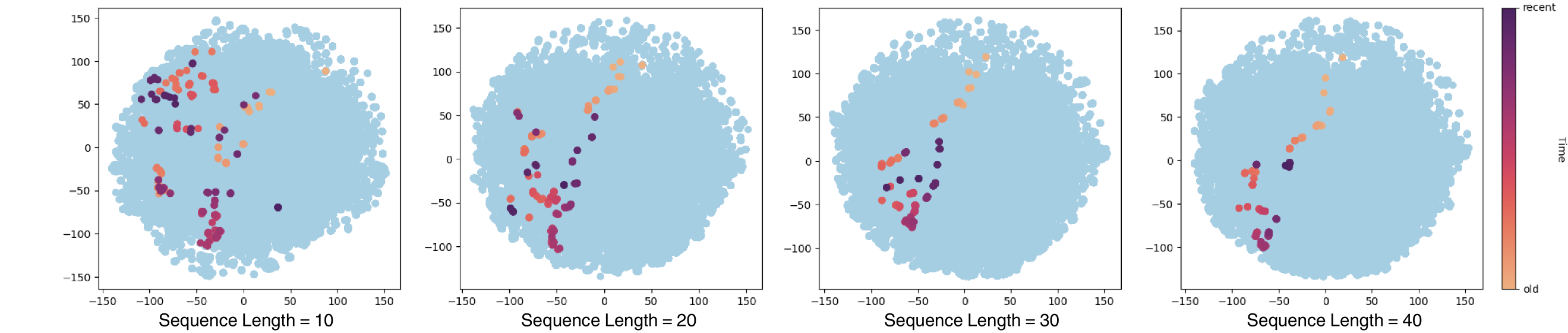}
    \caption{t-SNE visualization of the \texttt{Mistral 7B} embedding space for difference sequence lengths. The evolution of the same student's embedding through time is highlighted.}
    \Description{Four side-by-side t-SNE visualizations showing the Mistral 7B embedding space at different sequence lengths (10, 20, 30, 40). Each plot features a light blue circular cluster of points, with the evolution of a single student's embedding through time highlighted in colors ranging from orange (old) to purple (recent).}
    \label{fig:tsne-mistral}
\end{figure*}

\subsection{MCQBert: Correct Answer Prediction}
\label{sec:mcqbert}
This section introduces \texttt{MCQBert}, a model developed to predict students' responses to MCQs using only the question text\footnote{\texttt{MCQBert} is available at \url{https://huggingface.co/collections/epfl-ml4ed/student-answer-forecasting-edm-2024-663b7c20bb2aa3273dda4de2}}. The downstream task and primary objective of \texttt{MCQBert} is to accurately identify the correct answer(s) from the given options in each MCQ in the dataset.  The architectural design of \texttt{MCQBert} is illustrated in Figure \ref{fig:arch-bert}. It represents the \texttt{LernnaviBERT} model's application in processing both the question and a potential answer as input sequences. The transformer encoder component of \texttt{LernnaviBERT} processes the sequences before it is passed to a binary classifier, which outputs `1' for a correct answer and `0' for an incorrect one. 

\vspace{1mm} \noindent {\textbf{Data Split}}.
To formulate the MCQ prediction challenge as a binary classification task, each MCQ datapoint is decomposed into separate instances, each pairing the question with a possible answer option. The model thus aims to assign a `1' to a correct or student-selected answer and a `0' to an incorrect or unselected answer. 

We implement a partitioning ratio of 80/10/10 for training, validation, and testing, respectively. Notably, each MCQ occurs multiple times within the dataset, corresponding to different answers. To rigorously assess the model's ability to generalize and accurately answer new MCQs, we ensure that individual MCQs are exclusively allocated to either the training or the testing set. In other words, all occurrences of a particular MCQ are confined to a single subset.

\noindent {\textbf{Experiments}.
To evaluate the performance of our models in predicting correct answers to MCQs, two distinct experiments were conducted.
In the first one, the model was fine-tuned using the designated training set, and its generalization capacity was assessed on a separate test set comprising unseen questions, validating its ability to respond to MCQs beyond the training data.
The second experiment involved training the model on the entire MCQ dataset, confirming its effective learning of correct answers across all MCQs, ensuring that any failure in predicting a student's response did not arise from a lack of knowledge of the correct answer.
In the final phase of our pipeline for predicting student responses to MCQs, the model trained on the complete dataset was fine-tuned, allowing it to utilize its comprehensive knowledge of the correct answers when making predictions.

\label{sec:correct-answer}

This section describes the evaluation of \texttt{MCQBert}'s performance in the specific task of predicting correct answers to MCQs. The evaluation consisted of two distinct experiments, each employing a different training procedure to assess the model's predictive capabilities.

\noindent {\textbf{Experiment 1: Model Evaluation Against Unseen MCQs}}. The first experiment aimed to evaluate the ability of the model to predict the correct answers of previously unseen MCQs accurately. The models were fine-tuned for one epoch on a designated training set and subsequently evaluated on a separate test set. The evaluation metrics included MCC, F1 score, and accuracy.

The results, as summarized in Table \ref{tab:results-exp1}, contrast the performance of \texttt{MCQBert} with that of a Dummy Classifier, a baseline always predicting the majority class (i.e. 0). This comparison is useful for evaluating the effectiveness of \texttt{MCQBert} beyond simple chance or biased class distribution.

The performance metrics indicate that \texttt{MCQBert} outperforms the baseline Dummy Classifier, evidencing its capability to discern correct answers in the context of MCQs. 

\noindent {\textbf{Experiment 2: Model MCQs Retention Evaluation}}. The second experiment was designed to evaluate \texttt{MCQBert}'s capacity for retaining correct answers after being fine-tuned on the entire \textit{Lernnavi} MCQ dataset. The model was then tested on the same dataset to assess its ability to recall the correct answers, effectively evaluating its memorization capability. Not surprisingly, \texttt{MCQBert} achieved an MCC of 0.983, indicating nearly perfect recall of the correct answers within the dataset. This high level of performance is further corroborated by F1 scores of 0.993 for class 0 and 0.989 for class 1. The accuracy score of 0.992 reinforces the model's strong predictive capability and suggests that \texttt{MCQBert} model has learnt the correct answers to the vast majority of the MCQs present in the dataset, and we can therefore exploit this knowledge in the next step.

\subsection{Reproducibility for MCQStudentBERT}
\label{AppendixStudent}

 Similar to the MCQ prediction task without student data, we split the dataset into training, validation, and test datasets using an 80/10/10 split. In contrast to the base model (\texttt{MCQBert}), where each question-answer pair was exclusively assigned to one subset, in this task, individual MCQs can be present in both training and testing phases due to multiple representations within the dataset. This decision allows the model to leverage prior history related to specific questions, enhancing predictive accuracy by considering responses from students who share similar characteristics with the target student.

\noindent {\textbf{Evaluation Metrics}}. We assess our models using three different metrics: the \textit{Matthews Correlation Coefficient} (MCC) for binary classification, the \textit{F1 score} for balancing precision and recall, and the \textit{accuracy score} for overall predictions. MCC and F1-score are effective even if the classes are strongly imbalanced, motivating our evaluation choices. While accuracy and F1-score range between 0 to 1, MCC is a correlation coefficient value between -1 and +1 (+1: perfect prediction, 0: average random prediction, -1: inverse prediction). 

\subsection{Impact of sequence length on latent space representations}
We examined the sequence length impact on the \texttt{Mistral 7B} embeddings. Similar to Figure \ref{fig:tsne}, Figure \ref{fig:tsne-mistral} shows a single student's embeddings across varying sequence lengths. Echoing the behavior of the \texttt{LernnaviBERT} embedding, we note a discernible diagonal progression in the embeddings when the sequence length is greater than 10. The trend suggests that as the sequence length increases, the student's representation in the embedding space demonstrates a more pronounced diagonal trajectory, transitioning methodically from older to more recent embeddings with a progressively smoother evolution.

\end{document}